\documentclass[sigconf, nonacm, screen]{acmart}

\usepackage{algorithm}
\usepackage{algpseudocode}
\usepackage{tcolorbox}
\tcbuselibrary{listings,breakable}
\usepackage{listings}
\lstdefinestyle{trace}{
  basicstyle=\ttfamily\footnotesize,
  columns=fixed,
  keepspaces=true,
  upquote=true,
  breaklines=true,
  breakatwhitespace=false,
  showstringspaces=false,
  tabsize=2,
  captionpos=b
}
\renewcommand\footnotetextcopyrightpermission[1]{}
\title{Molecular Lead Optimization via Agentic Tool Planning}

\author{Lingxiao Li}
\affiliation{%
  \institution{University of Michigan}
  \city{Ann Arbor}
  \state{Michigan}
  \country{USA}
}
\email{llingxia@umich.edu}

\author{Haobo Zhang}
\affiliation{%
  \institution{University of Michigan}
  \city{Ann Arbor}
  \state{Michigan}
  \country{USA}
}
\email{haobozha@umich.edu}

\author{Ruohao Fan}
\affiliation{%
  \department{Department of Biochemistry and Molecular Medicine}
  \institution{University of California, Davis}
  \city{Davis}
  \state{California}
  \country{USA}
}
\email{rhfan@health.ucdavis.edu}

\author{Bin Chen}
\affiliation{%
  \department{Department of Pharmacology and Toxicology}
  \institution{Michigan State University}
  \city{East Lansing}
  \state{Michigan}
  \country{USA}
}
\email{chenbi12@msu.edu}

\author{Jiayu Zhou}
\affiliation{%
  \institution{University of Michigan}
  \city{Ann Arbor}
  \state{Michigan}
  \country{USA}
}
\email{jiayuz@umich.edu}

\begin{document}

\begin{abstract}

Drug discovery is a lengthy and resource-intensive process composed of multiple stages.
Among these stages, lead optimization plays a critical role in transforming
early hit compounds into viable drug candidates.
This stage requires improving ADMET-related properties through subtle structural
refinement while preserving key molecular substructures responsible for binding
affinity to disease targets.
Recent advances in artificial intelligence have shown promise in accelerating various aspects of drug discovery; however, most existing approaches to lead optimization rely on one-step molecular optimization, which fail to account for the long-term consequences of sequential design decisions. To address this limitation, we propose TRACE, a trajectory-aware, LLM-reasoning agent for molecular lead optimization that formulates tool selection as a sequential decision-making problem over action trajectories. Given a lead molecule and an optimization objective, TRACE makes trajectory-aware decisions over molecular optimization tools, enabling forward-looking refinement under structural constraints. Experiments on multiple ADMET optimization tasks show that our agent achieves higher optimization success, larger property improvements, and higher validity, while preserving molecular similarity compared to baseline models.

\end{abstract}

\maketitle

\section{Introduction}
\label{sec:introduction}

Drug discovery is a lengthy, costly, and high-risk process that spans multiple stages~\citep{paul2010improve}, including target identification, hit discovery, molecular lead optimization, and downstream preclinical and clinical evaluation~\citep{jorgensen2009efficient}.
Among these stages, molecular lead optimization plays a particularly
critical role in refining lead molecules into drug-like candidates suitable for clinical development~\citep{keseru2009influence}.
At this stage, candidate molecules must be optimized to satisfy a broad range of
ADMET-related requirements while preserving core substructures responsible for
binding affinity and target engagement through \emph{small, localized modifications}.
Because these final molecules are subsequently evaluated in preclinical studies, failures to meet desired objectives can have outsized scientific and financial consequences, as development costs increase dramatically once a compound progresses further down the pipeline~\citep{waring2015analysis}.

Despite its importance, lead optimization remains exceptionally challenging.
Improving desired properties through only small, localized structural modifications
is difficult, as such changes must navigate a tightly constrained
chemical space to achieve specific objectives while preserving other critical properties.
Although each objective is guided by empirical rules,  
 even experienced medicinal chemists often rely on extensive trial
and error to gradually refine candidates that meet all the criteria.
This process is time-consuming and difficult to scale, making lead optimization
a major bottleneck in the early discovery stage~\citep{hughes2011principles}.

Recent advances in artificial intelligence offer new opportunities to transform
modern drug discovery workflows~\citep{mialon2023augmented, schick2023toolformer}.
In particular, large language models (LLMs) have emerged as powerful agentic
controllers that can plan multi-step workflows, reason over intermediate results,
and interface with domain-specific computational tools.
Compared to independent, standalone models, agentic systems can dynamically
select and coordinate heterogeneous tools in response to the current optimization
context. A growing body of work has demonstrated the potential of such tool-using LLM
agents across different stages of drug discovery.
For example, ChemCrow augments an LLM with a collection of expert-designed
chemistry tools into a unified interface, enabling autonomous planning and execution for drug design~\citep{m2024augmenting}.
Similarly, CACTUS combines an LLM with standard cheminformatics tools to enable
molecular property assessments~\citep{mcnaughton2405cactus}.
DrugAgent applies agentic LLMs to drug discovery by coordinating multiple
specialized components such as high-throughput screening and drug-target interaction modules~\citep{inoue2025drugagent, hertzberg2000high, yamanishi2008prediction}.
At a broader workflow level, LIDDIA demonstrates that an LLM-based planner can coordinate multiple stages of in silico drug discovery, including generation,
screening, and optimization~\citep{averly2025liddia}.
These systems demonstrate that LLM-driven agents can effectively
orchestrate diverse tools and adapt strategies on the fly, a capability that is well suited to the iterative and decision-intensive nature of modern drug discovery pipelines.

Despite these advances, most chemistry agents focus on tool-augmented reasoning or end-to-end design, and do not explicitly model \emph{lead-conditioned optimization} as a dedicated stage.
Even when an optimization component is included, it is formulated as a population-level generic mutation within a broader pipeline rather than an explicit, per-lead constrained refinement process~\citep{jensen2019graph}.
To complement prior work and characterize this missing setting, we evaluate representative standalone lead-optimization baselines. However, even strong baselines can be brittle on a per-molecule basis: they may generate invalid structures, yield negligible
improvement on the target property, or violate structural constraints, indicating that no single optimization model is consistently reliable across molecules and objectives.
This motivates a tool-diverse, decision-making formulation, where multiple
functionally related but behaviorally distinct optimizers are coordinated to
improve robustness and optimization quality. However, naively executing multiple tools in parallel at each optimization step
incurs prohibitive computational cost.
This tension between optimization effectiveness and resource efficiency motivates
the need for agentic strategies that can selectively coordinate multiple tools,
rather than relying on exhaustive parallel execution.

To address these limitations, we introduce \textsc{TRACE} (\textbf{TR}ajectory-\textbf{A}djusted \textbf{C}hemical \textbf{E}diting), a trajectory-aware LLM-reasoning agent for molecular lead optimization that formulates tool selection as sequential decision-making over action trajectories under structural constraints that explicitly balances effectiveness, robustness, and computational efficiency.
\textbf{Our contributions} span both machine learning (ML) methodology and practical drug discovery needs.
From a \emph{ML} perspective, we introduce: (i) an \emph{LLM-guided lead-optimization agent} that coordinates a heterogeneous set of optimization tools under structural constraints, (ii) an \emph{in-context self-correction} mechanism that refines tool instructions from execution-time failures to stabilize optimizers, (iii) an \emph{anchored multi-step evolutionary exploration} procedure that progressively accumulates favorable modifications, and (iv) a \emph{similarity-guided trajectory reuse} strategy that retrieves and continues effective tool-usage trajectories from historical experience to reduce the need for exhaustive parallel tool execution.
From a \emph{drug discovery} perspective, our \textsc{TRACE} achieves high validity and success under
similarity constraints while delivering large property improvements across multiple ADMET optimization tasks.

\section{Related Work}
\label{sec:related_work}

Recent advances in large language models have catalyzed a shift from
standalone models toward \emph{agentic} systems that can decompose complex tasks into sequences of tool executions and iteratively adapt their decisions using accumulated context.
This paradigm has demonstrated strong potential in domains where effective
problem solving requires coordinating heterogeneous capabilities, including
software engineering and scientific workflows.

In drug discovery, recent works have explored tool-augmented LLM agents.
ChemCrow equips an LLM with chemistry-specific tools, enabling the model to access external knowledge and
autonomously plan and execute multi-step workflows on drug discovery~\citep{m2024augmenting}.
CACTUS follows a similar philosophy by integrating standard cheminformatics tools
with an LLM to support molecule property evaluations~\citep{mcnaughton2405cactus}.
DrugAgent applies agentic LLMs and coordinates multiple specialized components for ADMET prediction, high-throughput
screening, and drug-target interaction prediction~\citep{inoue2025drugagent}.
At a broader workflow level, LIDDIA demonstrates that an LLM-based planner can coordinate multiple stages of the drug discovery pipeline, including generation, screening, and optimization into an end-to-end agentic system~\citep{averly2025liddia}. Together, these works highlight the promise of LLM-driven agents as flexible controllers for complex, multi-stage discovery pipelines. However, existing agentic systems rarely target lead-conditioned optimization under explicit structural constraints.
Even when an optimization module is included~\citep{averly2025liddia}, it is implemented as a population-level mutation rather than a local refinement procedure centered on a given lead~\citep{jensen2019graph}.

\section{Background and Task Definition}
\label{sec:background_task}

\paragraph{Lead Optimization as a Constrained Problem.} 
In drug discovery, lead optimization refers to the process of improving a candidate molecule’s pharmacologically relevant properties through targeted chemical modifications, while preserving the core structural features of the
original lead~\citep{bemis1996properties}.
Unlike unconstrained molecular generation, lead optimization must operate within a highly restricted chemical space: only localized edits are permitted, and large scaffold changes are generally undesirable.
As a result, lead optimization is challenging, as small structural
changes may induce complex and conflicting effects on molecular properties.

A central objective of lead optimization is the improvement of ADMET properties, which characterize a molecule’s absorption, distribution, metabolism, toxicity, and overall drug-likeness~\citep{veber2002molecular}.
Overall, lead optimization is best viewed as a constrained optimization
problem, where property improvement must be balanced against strict structural and
chemical feasibility requirements~\citep{jensen2019graph, dey2025gellm3o}.

\vspace{+0.05in}
\paragraph{Task Formulation.} 
In this work, we study lead optimization through a set of single-property
optimization tasks that serve as representative benchmarks.
Each task is defined by an input molecule $m_0$ and a target ADMET property to be improved~\citep{zhu2023sample}.
The objective is to generate a chemically valid molecule $m$ that achieves a
higher value of the target property (or a lower value in the case of toxicity) while remaining structurally similar to the input lead.

To enforce structural fidelity, we impose an explicit similarity constraint between the input and output molecules.
This constraint ensures that optimization is achieved through minimal, localized
edits rather than extensive structural rewiring, reflecting practical lead optimization settings where the original scaffold is largely preserved.
Specifically, structural similarity is measured using Morgan
fingerprint–based Tanimoto similarity~\citep{todeschini2008handbook}, and only molecules with
$\mathrm{SIM} \ge 0.5$ are considered valid optimization outcomes.

\vspace{+0.05in}
\paragraph{Target Properties.} 
We consider five ADMET-related endpoints that are commonly used to evaluate lead optimization quality.
\textsc{pLogP} characterizes molecular lipophilicity and influences absorption and permeability.
\textsc{QED} provides an aggregate measure of drug-likeness.
\textsc{HIA} (human intestinal absorption) reflects the likelihood of oral absorption.
\textsc{BBBP} (blood--brain barrier permeability) captures distribution behavior, particularly for central nervous system targets.
\textsc{Mutagenicity} is a toxicity-related endpoint, where lower values indicate reduced risk. Together, these properties span absorption, distribution, toxicity, and overall
drug-likeness~\citep{lipinski2012experimental}. For each task, higher values are preferred except for mutagenicity, where the optimization objective is to reduce risk.
The task can therefore be summarized as follows: given an input molecule, generate a structurally similar analog that improves a specified ADMET property while respecting chemical validity and structural constraints.

\section{Motivation}
\label{sec:motivation}

Lead optimization is a critical stage of drug discovery that requires improving
molecular properties under strict structural constraints.
Despite recent progress in AI-assisted molecular design, effectively automating
this process remains challenging. Motivated by these challenges, we analyze empirical behaviors of existing lead optimization tools and identify several key observations that collectively motivate a multi-tool, trajectory-aware agentic framework.

\begin{figure}[htbp]
  \centering
  \includegraphics[width=1.0\linewidth]{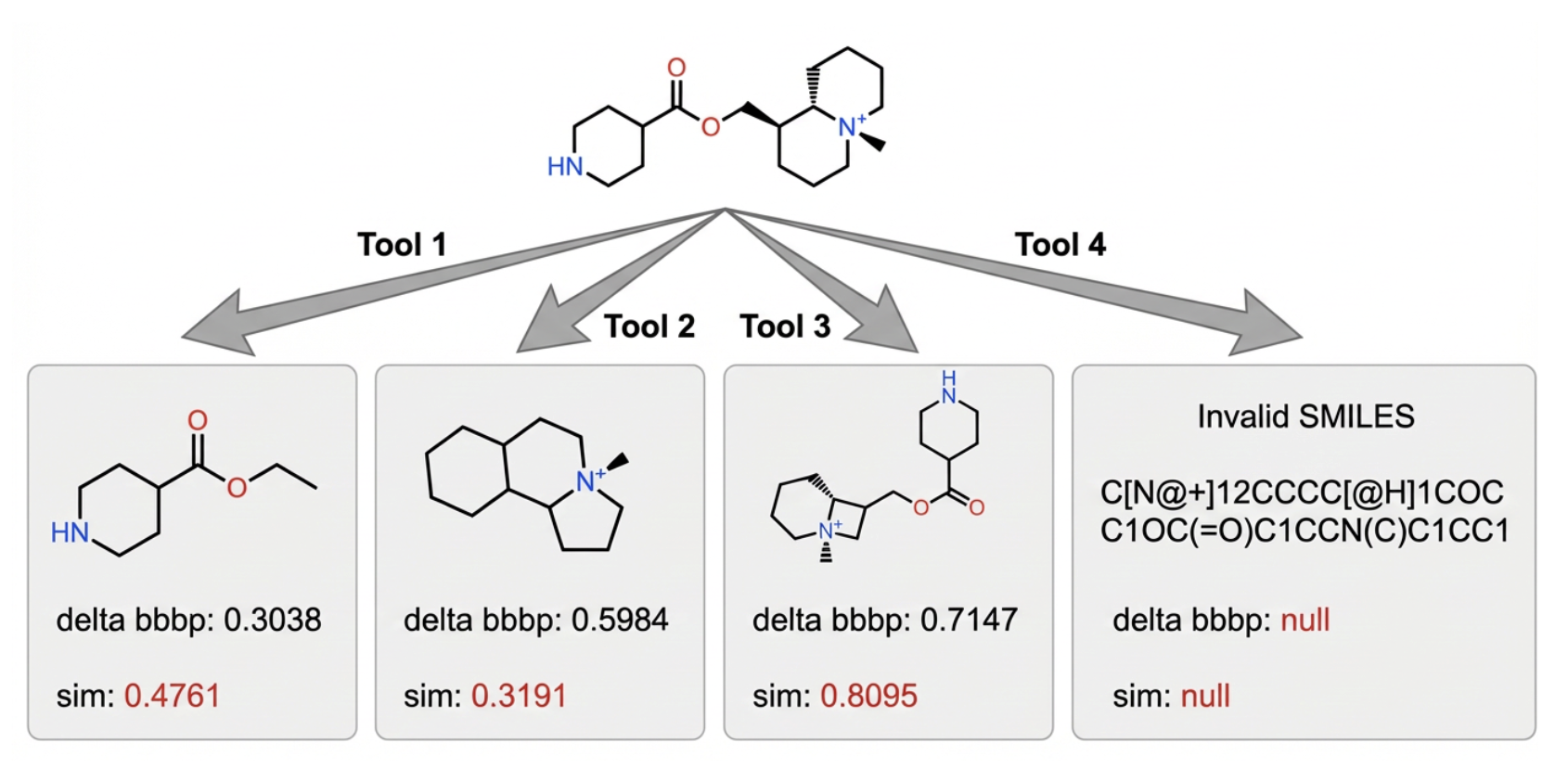}
  \vspace{-0.2in}
  \caption{Tool heterogeneity in lead optimization. Given the same lead molecule (top), different optimization tools propose qualitatively different edits and outcomes.}
  \label{fig:tool_heterogeneity}
  \vspace{-0.2in}
\end{figure}

\vspace{+0.05in}
\paragraph{Tool Heterogeneity.}
Our first observation is that different lead optimization tools~\citep{dey2025gellm3o}, even when achieving comparable performance on aggregate benchmark metrics, often behave very differently at the level of individual molecules (Figure~\ref{fig:tool_heterogeneity}).
Given the same lead compound, distinct tools may propose different
structural modifications: some introduce localized functional-group changes,
others drift toward larger scaffold alterations, and some fail to produce valid
chemical structures altogether.
These differences are frequently obscured when performance is averaged across
large test sets, but become evident when examining per-instance outcomes, such as property improvement, structural similarity, and chemical validity.

This per-molecule heterogeneity reveals an important limitation of single-tool
optimization.
No single optimizer is consistently reliable across leads and objectives: a tool that succeeds on one molecule may fail or violate constraints on another.
Consequently, committing to a single optimization strategy can reduce robustness and lead to missed optimization opportunities.
Instead, effective lead optimization should leverage the complementary strengths of multiple tools, allowing different behaviors to be exploited when appropriate.
This observation naturally motivates an agentic formulation, where a controller
can adaptively select among heterogeneous lead optimization tools based on the current molecular context, rather than relying on any single optimizer.

\vspace{+0.05in}
\paragraph{Feedback Enables Self-Correction.}
Beyond tool diversity, we observe that \emph{execution-time feedback} plays a crucial role in improving lead optimization outcomes.
When a lead optimization tool produces an unsatisfactory result, such as an invalid molecule, a constraint violation, or no improvement on the target property, the system records the failure, which helps avoid repeating similar ineffective behaviors in subsequent attempts.

\begin{figure}[htbp]
  \centering
  \includegraphics[width=1.0\linewidth]{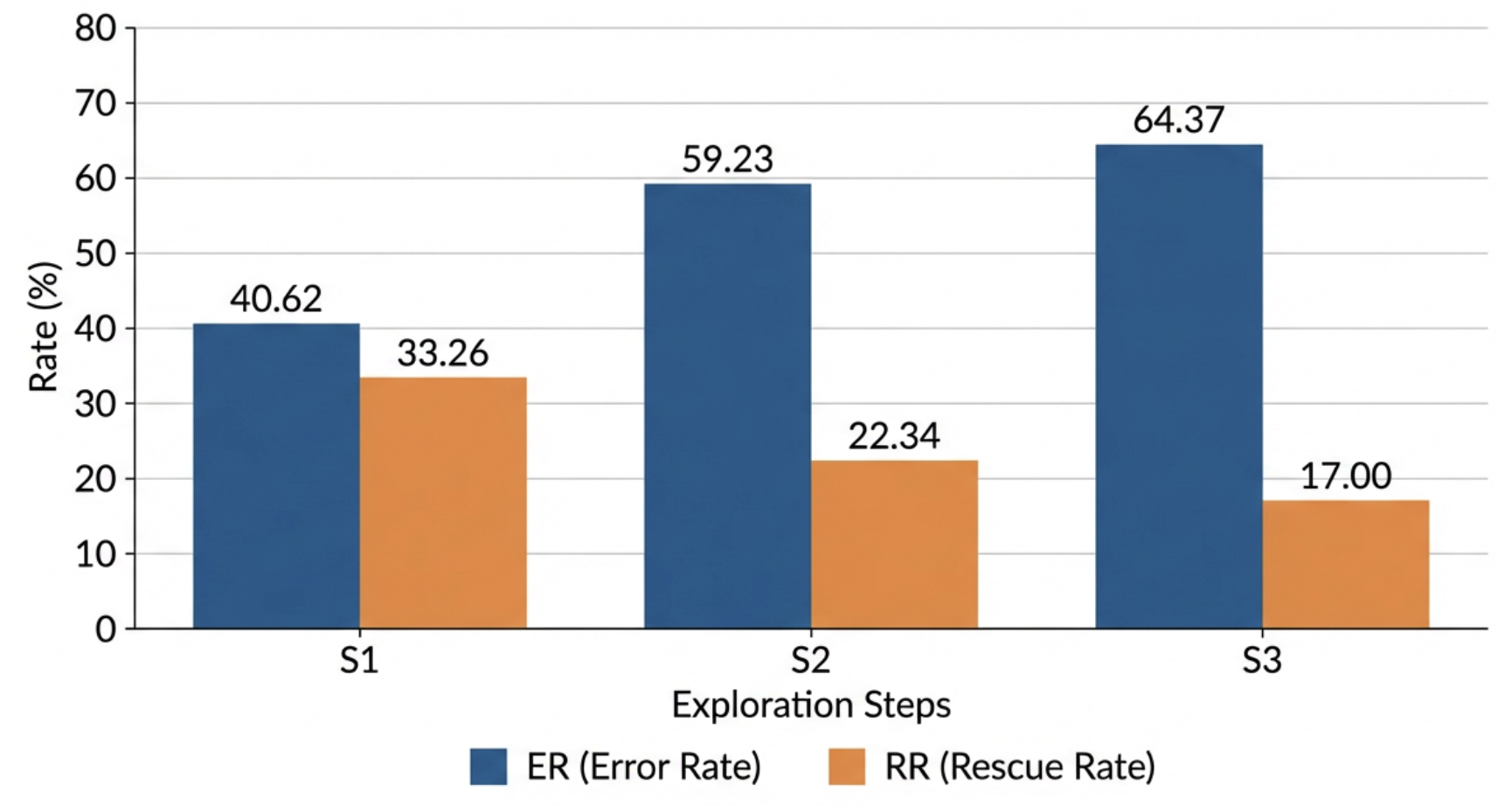}
\vspace{-0.2in}
\caption{In-context self-correction across TRACE exploration steps on BBBP.
Error Rate denotes the proportion of generated molecules that fail to satisfy the optimization objective among all generated candidates at current step. Rescue Rate measures the fraction of failed generations that can be corrected into successful candidates through subsequent self-correction steps.}
\label{fig:self_correction_motivation}
\vspace{-0.3in}
\end{figure}

Importantly, this form of self-correction is \emph{specific to LLM-based optimization tools}.
Because these tools generate by conditioning on natural-language instructions, they can directly leverage execution-time feedback, such as failed examples or refined constraints provided at inference time.
By augmenting the original prompt with contextualized failure information, the same model can adjust its generation behavior through reasoning, without any parameter updates or retraining.
As illustrated in Figure~\ref{fig:self_correction_motivation}, this instruction refinement recovers a substantial fraction of candidates that would otherwise fail, effectively turning prior mistakes into guidance for subsequent generation.
Overall, execution-time feedback provides a practical mechanism for stabilizing LLM-based lead optimization, motivating our explicit integration of in-context self-correction into the TRACE’s optimization steps.

\vspace{+0.05in}
\paragraph{Budget-Aware Multi-Step Exploration.}
While in-context self-correction improves the reliability of individual tool
executions, it does not fundamentally expand the optimization capacity of the
underlying models.
From a search perspective, most lead optimization tools operate as
\emph{single-step} procedures guided by frozen model parameters, which limits the
magnitude of improvement achievable in any one application.
As a result, even when a tool succeeds, the resulting modification is often small
and insufficient to satisfy challenging property objectives.

\begin{figure}[htbp]
  \centering
  \includegraphics[width=1.0\linewidth]{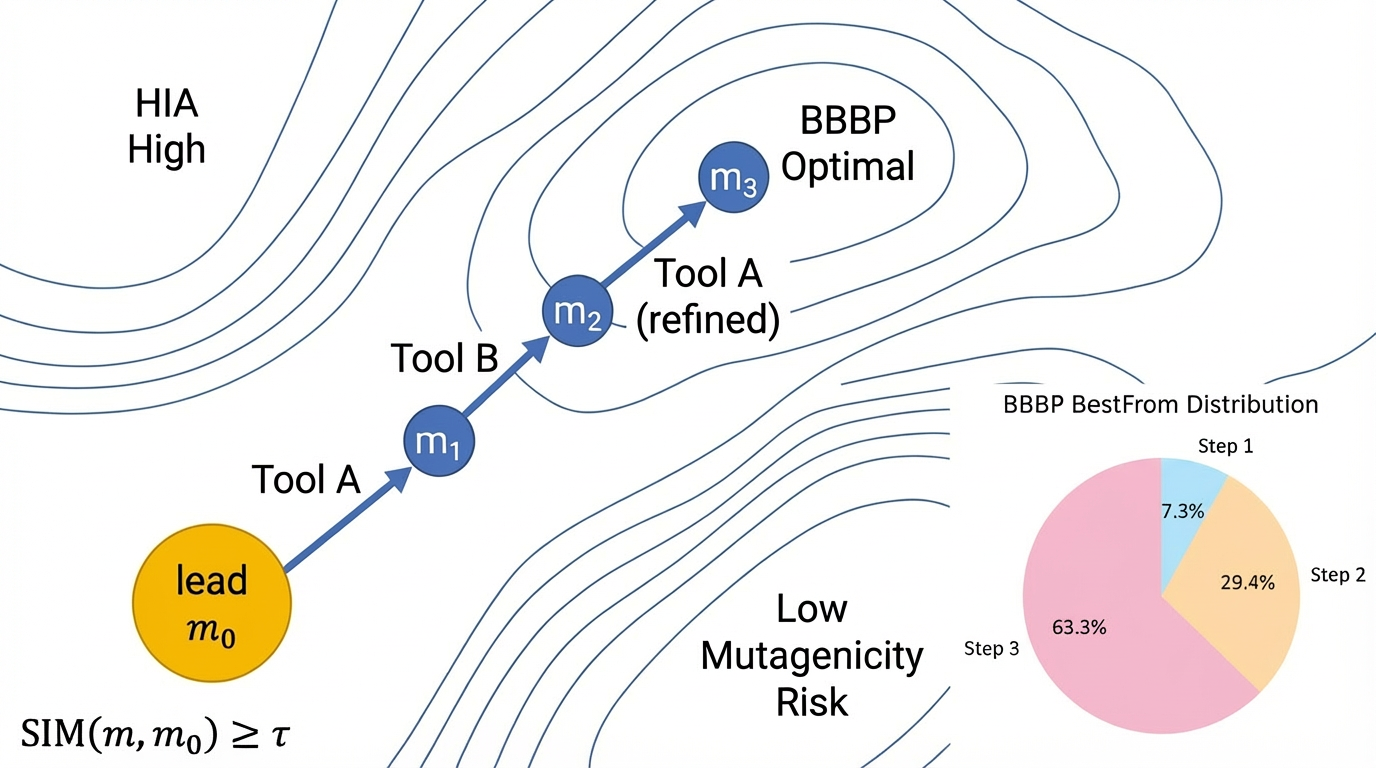}
  \caption{\textbf{Multi-step exploration under similarity constraints.}
Starting from lead $m_0$ with $\mathrm{SIM}(m,m_0)\ge \tau$, additional optimization with parallel multi-tool executions often uncover better candidates than a single step (right: \textsc{BestFrom} by step)}
  \label{fig:space_searching}
\end{figure}

Figure~\ref{fig:space_searching} illustrates this phenomenon from a local search perspective.
We observe that while the improvement achievable by a single optimization step is inherently limited, \emph{multi-step exploration} often yields substantially better optimization outcomes by discovering stronger candidates at later steps.
In practice, many of the best-performing molecules emerge only after additional refinement beyond the initial optimization step.
However, achieving these gains typically requires applying multiple optimization tools in parallel at each step in order to explore diverse modification directions, which incurs significant computational cost.
This creates a fundamental tension: deeper exploration is highly beneficial for constrained lead optimization, yet executing many tools at every step quickly becomes infeasible under realistic resource budgets.

\begin{figure}[htbp]
  \centering
  \includegraphics[width=1.0\linewidth]{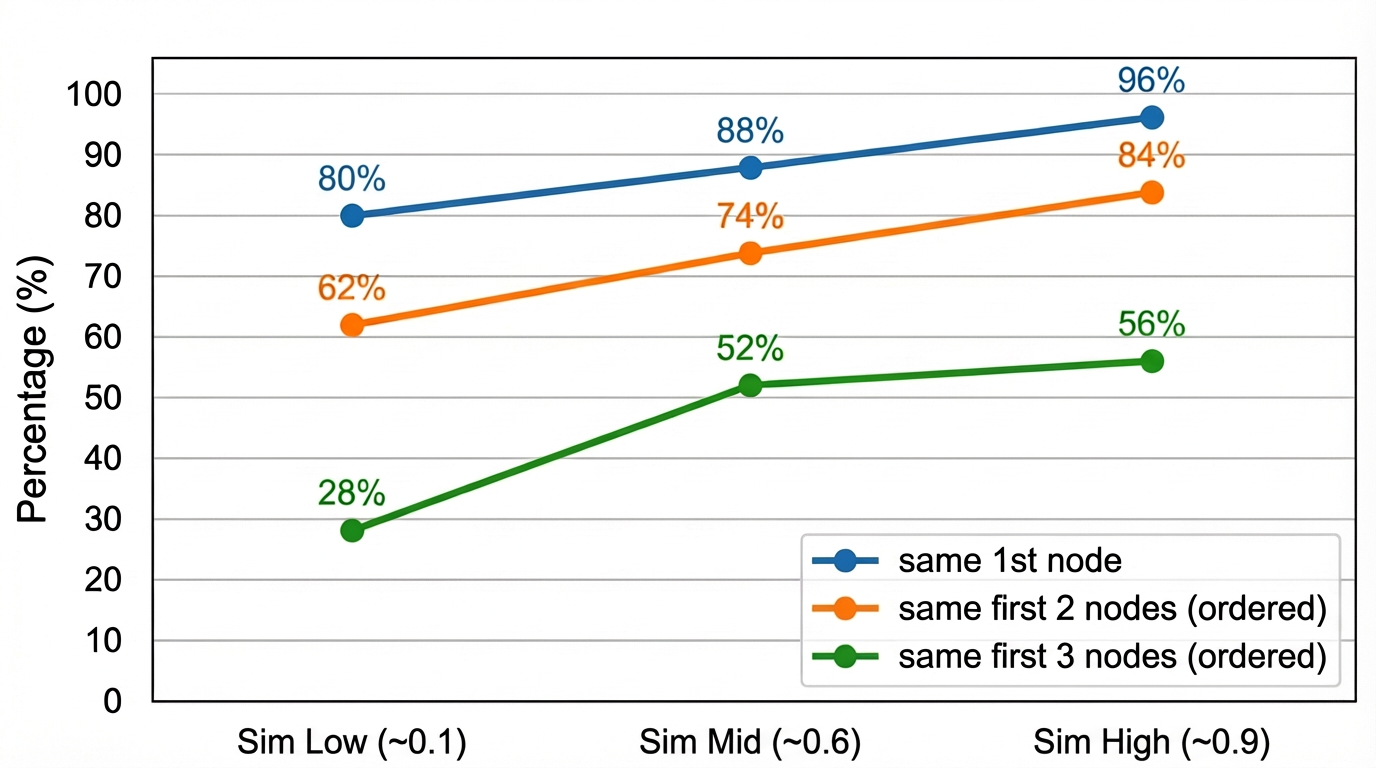}
  \caption{\textbf{Trajectory similarity correlates with structural similarity.} Across similarity regimes, paired molecule groups with higher average structural similarity are more likely to share the same optimization trajectory (matching the first $k$ tool-action nodes).}
  \label{fig:trajectory_sim}
\end{figure}

To address this tension, we observe that effective optimization behaviors are often \emph{trajectory-dependent} rather than action-dependent.
Empirically, as shown in Figure~\ref{fig:trajectory_sim}, structurally similar molecules tend to share similar and effective tool-usage sequences, indicating that historical optimization trajectories can be reused and adapted.
This motivates a trajectory-aware strategy in which the agent retrieves a
promising tool-usage template from prior experience and executes it sequentially, \emph{one tool per step}, instead of exhaustively evaluating all tools in parallel. By reusing and adapting trajectory-level knowledge, the agent can maintain strong optimization performance while dramatically reducing computational cost.
Importantly, this mechanism operates entirely through offline experience and
in-context conditioning, requiring no additional training or parameter updates.
As a result, the agent can efficiently generalize to unseen molecules, new
property objectives, and evolving tool sets, even under tight resource
constraints.

\section{Method}
\label{sec:method}

Motivated by the observations in Section~\ref{sec:motivation}, we propose TRACE, an agentic framework for molecular lead optimization that explicitly treats
optimization as a sequential, decision-driven process under structural
constraints.
Our design addresses three key challenges revealed by prior analysis:
(i) the heterogeneous and complementary behaviors of existing lead optimization
tools, (ii) the need to incorporate execution-time feedback to improve reliability,
and (iii) the tension between effective multi-step exploration and limited
computational budgets.

\begin{figure}[htbp]
  \centering
  \includegraphics[width=1.0\linewidth]{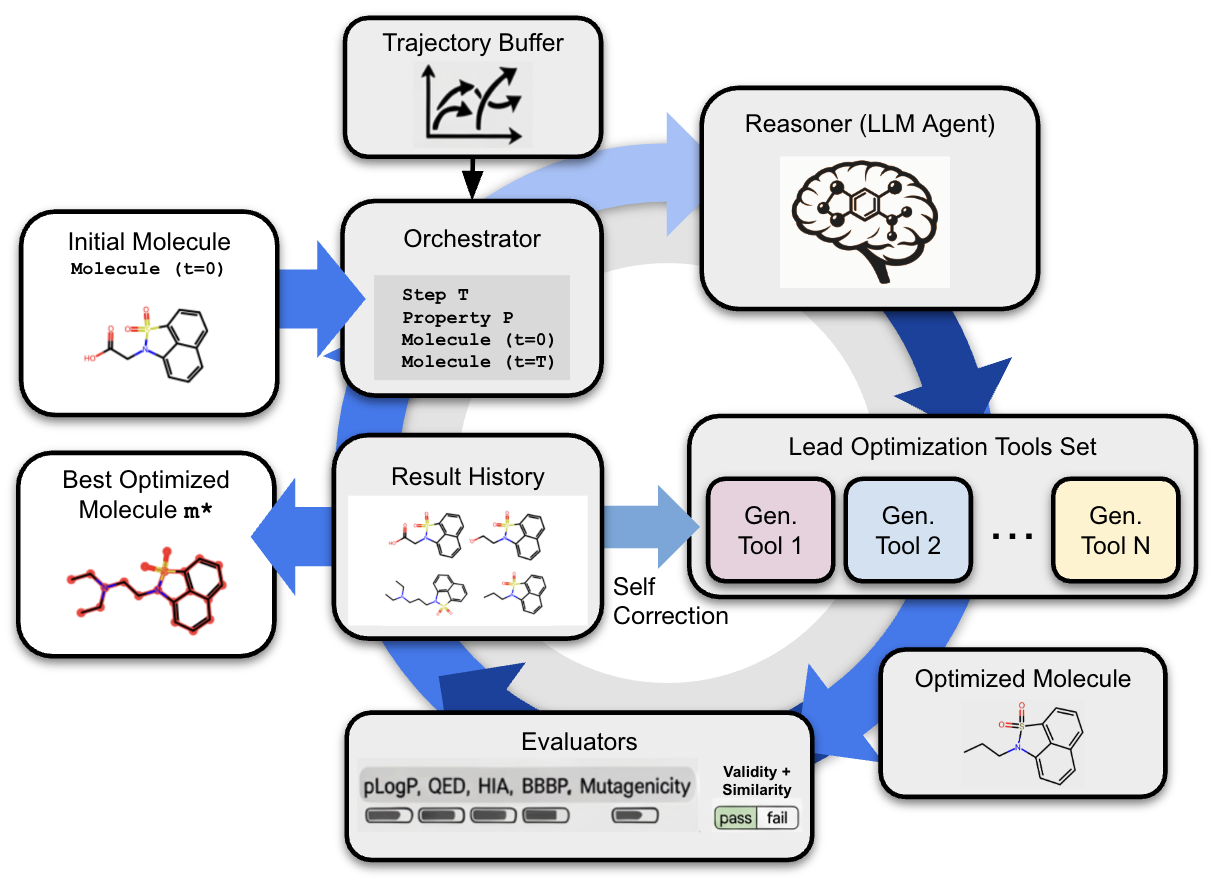}
  \caption{\textbf{TRACE overview.}
The orchestrator maintains the current lead state and routes context to an
LLM-based reasoner, which selects an optimization tool and issues instructions.
Lead optimization tools propose candidate molecules that are scored by predictive evaluators and filtered by different checks, with failed outcomes for self-correction and successful outcomes for additional explorations.
An offline trajectory buffer provides reusable optimization experience.}

\label{fig:overall_framework}
\end{figure}

\vspace{+0.05in}
\paragraph{Framework.}
At a high level, TRACE is an agent equipped with a set of heterogeneous LLM-based lead optimization tools and enables adaptive coordination among them,
rather than committing to any single optimizer. Algorithm~\ref{alg:trajectory_tool_sequencing} summarizes the overall procedure. Starting from an input lead molecule and a target property objective, TRACE retrieves relevant historical trajectories based on molecular similarity, selects tools
conditioned on the evolving context, and applies them to generate new candidate molecules.
Throughout the process, favorable structural modifications are inherited and accumulated across steps, enabling gradual property improvement without introducing abrupt structure changes.

Figure~\ref{fig:overall_framework} illustrates the overall architecture of the proposed framework. It consists of six interacting components: (i) an LLM-based reasoner for decision making, (ii) an orchestrator that governs execution and information flow, (iii) tool-instruction memory and result-history memory for persistent system knowledge, (iv) a trajectory buffer that stores action-level optimization experience, (v) a collection of generative tools, and (vi) predictive evaluators. These components are coordinated by the orchestrator, forming a closed-loop optimization system.

\begin{algorithm}[t]
\caption{TRACE (Resource-Aware Similarity-Guided Tool Planning)}
\label{alg:trajectory_tool_sequencing}
\begin{algorithmic}[1]
\Require Initial molecule $m_0$, target property $p$, trajectory buffer $\mathcal{B}$,
similarity threshold $\tau$, tool set $\mathcal{T}$, LLM reasoner $\pi$,
orchestrator $\mathcal{O}$, evaluators $\mathcal{E}$, per-step tool budget $K$
\Comment{$K\in\{1,|\mathcal{T}|\}$}
\State $m \gets m_0$
\State $\texttt{traj} \gets \emptyset$

\For{\textbf{each} step $t=1,\dots,T$}
    \If{$K>1$} \Comment{high budget: parallel tools}
        \State $\mathcal{I} \gets \pi(m,p)$ 
        \State $m \gets \mathcal{O}(\mathcal{I}, m, m_0, p, \mathcal{E})$
        \Comment{incl.\ self-correction}
        \State \textbf{continue}
    \EndIf

    \Comment{low budget: single tool}
    \State $m_{\mathrm{ref}} \gets \textsc{Top1Similar}(m, p, \mathcal{B})$
    \If{$\textsc{Sim}(m, m_{\mathrm{ref}}) \ge \tau$}
        \State $\texttt{traj} \gets \textsc{Trajectory}(m_{\mathrm{ref}})$
    \ElsIf{$\texttt{traj}$ is empty}
        \State $\texttt{traj} \gets \pi(m,p)$
    \EndIf
    \State $m \gets \mathcal{O}(\texttt{traj}, m, m_0, p, \mathcal{E})$
    \Comment{incl.\ self-correction}
\EndFor
\end{algorithmic}
\end{algorithm}

\vspace{+0.05in}
\paragraph{In-Context Self-Correction.}
To improve execution robustness within a single optimization step, we introduce
\emph{in-context self-correction} for LLM-based lead
optimization tools. Self-correction operates entirely at inference time by
refining tool instructions based on immediate failure feedback, rather than
updating parameters or performing extensive search.

After invoking a tool, the agent detects failures such as invalid structures,
constraint violations, or lack of property improvement, then augments the
original prompt with the failed output and corrective guidance (Box~4.1) and
re-invokes the same tool. Self-correction is capped to one additional attempt
per step; if it still fails, the agent proceeds to the next exploration step.

\begin{tcolorbox}[
  title={Box 4.1: Self-Correction Prompt Template},
  label={box:self_correction_prompt},
  colback=black!1,
  colframe=black!35,
  fonttitle=\bfseries
]
\ttfamily\footnotesize
Modify the molecule to improve the target property by substituting functional
groups while keeping changes to the core minimal. Return the modified SMILES
wrapped in \texttt{<SMILES>...</SMILES>}.

\smallskip
{\color{red}\bfseries Before giving your final answer, think carefully. Here are poor solutions to avoid:}
{\color{red}\begin{itemize}\itemsep0pt
  \item \texttt{<FailCase1>}: invalid SMILES;
  \item \texttt{<FailCase2>}: \dots
\end{itemize}}
\end{tcolorbox}

\vspace{+0.05in}
\paragraph{Multi-Step Evolutionary Exploration.}
While self-correction stabilizes individual tool executions, it does not expand
the optimization capacity of the underlying tools, which typically act as
single-step editors with frozen parameters. To enable larger and more consistent
property improvements, we adopt a multi-step evolutionary exploration strategy that progressively refines candidates over time.

\begin{figure}[htbp]
  \centering
  \includegraphics[width=1.0\linewidth]{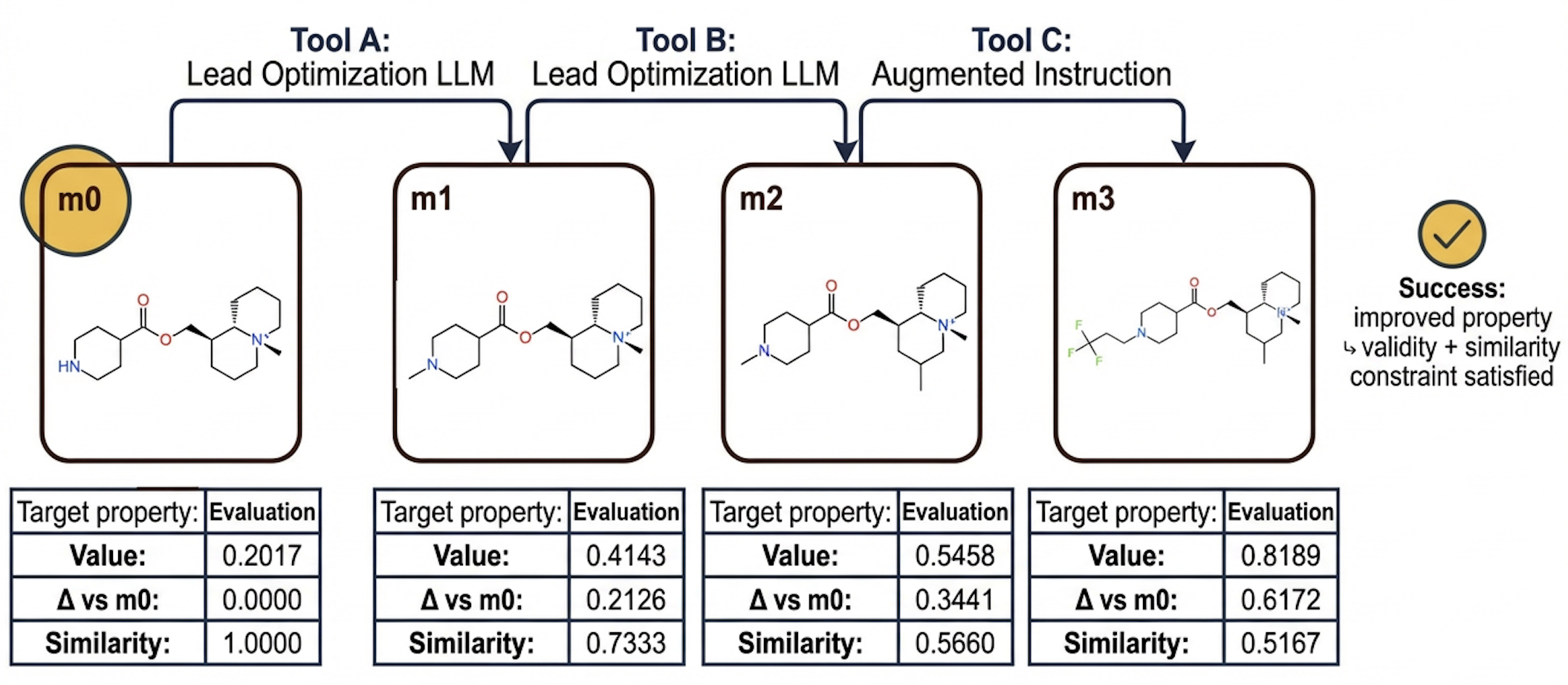}
  \caption{\textbf{Anchored multi-step exploration.}
Starting from lead $m_0$, the agent iteratively applies different optimization
tools to produce candidates ($m_1\!\rightarrow\!m_3$), selects the best molecule at
each step by improvement on the target property(e.g., BBBP), and uses it as the next-step starting point. All scores and constraints are evaluated relative to the original lead.}
  \label{fig:exploration}
  \vspace{-0.2in}
\end{figure}

At each exploration step, the agent evaluates all generated candidates with
predictive evaluators and selects the best-performing molecule based on relative
improvement on the target property. The selected molecule is then used as the
starting point for the next step, allowing favorable structural modifications to
be inherited and accumulated across iterations while avoiding abrupt changes that
could compromise key substructures. 
Crucially, all evaluations are anchored to the original lead molecule: structural similarity, property improvement, and constraint
satisfaction are always measured relative to this fixed reference. This anchoring
prevents uncontrolled drift and ensures that exploration remains a constrained
refinement process. The final output is
chosen as the best candidate observed across all steps (Figure~\ref{fig:exploration}).

\vspace{+0.05in}
\paragraph{Similarity-Guided Tool Trajectory Reuse.}
A key challenge in agentic lead optimization is trading off optimization quality
and computational cost. While executing all available tools in parallel at each
step can improve robustness, it quickly becomes infeasible as the tool set or
exploration depth grows. In realistic settings, the agent should be able to select only
one tool per step, yet still construct an effective multi-step trajectory.

We address this by introducing similarity-guided trajectory reuse for tool
sequencing. TRACE stores past optimization trajectories in an offline buffer,
where each trajectory records the tool sequence and the corresponding evaluation
outcomes. At execution time, TRACE computes structural similarity between the
current molecule and buffer entries. If a sufficiently similar reference is found,
TRACE retrieves a high-performing trajectory and follows it as a template for
subsequent steps; otherwise it continues the current trajectory or plans a new one
based on the current context. This retrieval mechanism enables multi-step planning
with a single tool call per step, requires no additional training, and adapts
naturally as objectives or available tools change under strict budgets.

\section{Experimental Settings}
\label{sec:experiments}

\begin{table*}[!ht]
\centering
\scriptsize
\setlength{\tabcolsep}{4pt}
\caption{Main results across five ADMET optimization tasks.}
\label{table:main}

\resizebox{\textwidth}{!}{\begin{tabular}{l*{20}{c}}
\hline
\textbf{Method} &
\multicolumn{4}{c}{\textbf{bbbp}} &
\multicolumn{4}{c}{\textbf{hia}} &
\multicolumn{4}{c}{\textbf{mutagenicity}} &
\multicolumn{4}{c}{\textbf{qed}} &
\multicolumn{4}{c}{\textbf{plogp}} \\
\cline{2-21}

& SR & SIM & RI & VR
& SR & SIM & RI & VR
& SR & SIM & RI & VR
& SR & SIM & RI & VR
& SR & SIM & RI & VR \\
\hline

\multicolumn{21}{c}
{\textbf{Molecule optimization tool}} \\
GraphGA &
\multicolumn{16}{c}{\textsc{n/a}} &
89.79 & 5.47 & 0.00 & 100.00 \\
\multicolumn{21}{c}{\textbf{General-purpose LLMs}} \\
Llama  & 30.69 & 50.94 & 33.03 & 63.36 & 33.33 & 51.30 & 73.70 & 71.79 & 34.06 & 43.57 & 15.63 & 71.97 & 23.02 & 43.73 & 15.60 & 65.46 & 28.57 & 60.77 & 64.44 & 71.42 \\
Mistral & 31.68 & 67.25 & 37.84 & 54.45 & 41.02 & 63.63 & 100.18 & 58.97 & 39.01 & 58.48 & 27.38 & 57.14 & 23.74 & 62.42 & 21.62 & 56.83 & 30.61 & 62.05 & 68.43 & 59.86\\
% GPT-4o  & & & & & & & & & & & & & & & & & & & & \\

\multicolumn{21}{c}
{\textbf{Foundational LLMs for Chemistry}} \\
ChemLLM & 11.88 & 0.11 & 0.00 & 11.88 & 12.82 & 0.00 & 0.00 & 12.82 & 6.59 & 1.21 & 0.00 & 12.08 & 2.15 & 0.00 & 0.00 & 11.51 & 0.00 & 0.00 & 0.00 & 2.72 \\
LlaSMol-CodeLlama & 89.10 & 23.38 & 99.63 & 98.01 & 64.10 & 40.12 & 89.36 & 97.43 & 72.52 & 25.30 & 31.76 & 98.90 & 67.62 & 25.09 & 45.01 & 98.56 & 29.25 & 35.76 & 46.65 & 99.31 \\
LlaSMol-Galactica & 85.14 & 24.42 & 45.38 & 99.00 & 55.26 & 39.13 & 108.60 & 100.00 & 73.07 & 24.62 & 16.60 & 97.80 & 54.67 & 18.12 & 35.98 & 97.84 & 30.61 & 44.07 & 30.94 & 99.31 \\
LlaSMol-Llama2 & 70.29 & 30.54 & 81.04 & 80.19 & 46.15 & 32.70 & 51.98 & 64.10 & 58.79 & 27.95 & 31.17 & 71.42 & 48.20 & 30.41 & 36.31 & 80.57 & 23.80 & 40.26 & 80.85 & 65.30 \\
LlaSMol-Mistral & 95.04 & 22.82 & 70.34 & 100.00 & 53.84 & 36.85 & 57.66 & 100.00 & 72.52 & 23.07 & 15.81 & 100.00 & 64.02 & 18.25 & 0.00 & 99.28 & 26.53 & 45.10 & 27.30 & 99.31 \\
        
\multicolumn{21}{c}{\textbf{Lead Optimization LLMs}} \\
GeLLMO-P(4)Llama   & 78.21 & 51.83 & 98.71 & 93.06 & 84.61 & 48.92 & 112.35 & 92.30 & 68.13 & 44.78 & 32.01 & 87.91 & 67.62 & 51.71 & 40.38 & 88.48 & 52.38 & 53.00 & 80.79 & 93.87 \\
GeLLMO-P(4)Mistral & 78.21 & 55.29 & 89.82 & 97.02 & 71.79 & 52.57 & 100.44 & 97.43 & 67.03 & 46.74 & 36.09 & 94.50 & 62.58 & 61.25 & 44.84 & 97.12 & 56.46 & 61.78 & 112.39 & 98.63\\
GeLLMO-P(6)Llama   & 81.18 & 50.48 & 88.63 & 91.08 & 74.35 & 38.04 & 102.78 & 92.30 & 73.62 & 40.19 & 45.12 & 92.30 & 72.66 & 48.80 & 40.91 & 92.08 & 48.29 & 53.28 & 109.72 & 89.79 \\
GeLLMO-P(6)Mistral & 74.25 & 58.85 & 69.74 & 98.01 & 79.48 & 47.35 & 106.10 & 97.43 & 78.57 & 49.46 & 39.78 & 95.05 & 61.87 & 61.28 & 36.94 & 94.96 & 59.18 & 63.60 & 113.51 & 93.87 \\
        
\multicolumn{21}{c}{\textbf{Lead Optimization Agents}} \\

TRACE-Online & 87.12 & 59.91 & 121.93 & 100.00 & 92.30 & 63.42 & 120.39 & 100.00 & 84.06 & 60.33 & 53.46 & 100.00 & 94.24 & 59.20 & 59.68 & 100.00 & 74.82 & 59.59 & 155.03 & 100.00 \\

TRACE-Retrieve & 97.02 & 59.48 & 136.23 & 100.00 & 97.43 & 60.67 & 128.68 & 100.00 & 92.30 & 59.59 & 54.64 & 100.00 & 96.40 & 59.40 & 60.61 & 100.00 & 89.11 & 61.19 & 163.05 & 100.00\\

TRACE-Parallel & 100.00 & 54.32 & 167.60 & 100.00 & 100.00 & 53.61 & 150.84 & 100.00 & 99.45 & 54.65 & 80.44 & 100.00 & 100.00 & 55.05 & 79.29 & 100.00 & 94.55 & 54.94 & 228.93 & 100.00\\
\hline

\end{tabular}}
\end{table*}

\paragraph{Baselines.} We compare TRACE against baselines from four categories: a classical molecule
optimization tool, general-purpose LLMs, chemistry-oriented foundation LLMs, and
lead-optimization LLMs. As a non-LLM reference, we include GraphGA~\citep{jensen2019graph},
which is also used as an optimization component in related agentic pipelines~\citep{averly2025liddia}.
We include representative general-purpose LLMs, Mistral-7B-Instruct-v0.3~\citep{mistral7b_instruct_v03}
and Llama-3.1-8B-Instruct~\citep{llama31_8b_instruct}, to assess how far generic
instruction-following can transfer to molecular optimization. We also evaluate
chemistry-oriented foundation LLMs pretrained or instruction-tuned on large-scale
chemical corpora, including ChemLLM~\citep{zhang2024chemllm} and LlaSMol variants
(LlaSMol-CodeLlama, LlaSMol-Galactica, LlaSMol-Llama2, and LlaSMol-Mistral)~\citep{yu2024llasmol}.
Finally, we compare against GeLLMo~\citep{dey2025gellm3o}, an instruction-guided
LLM designed for molecular lead optimization, which serves as a strong standalone
baseline for assessing the benefits of multi-tool coordination and trajectory-aware
decision making.

\begin{figure}[htbp]
  \centering
  \includegraphics[width=\linewidth]{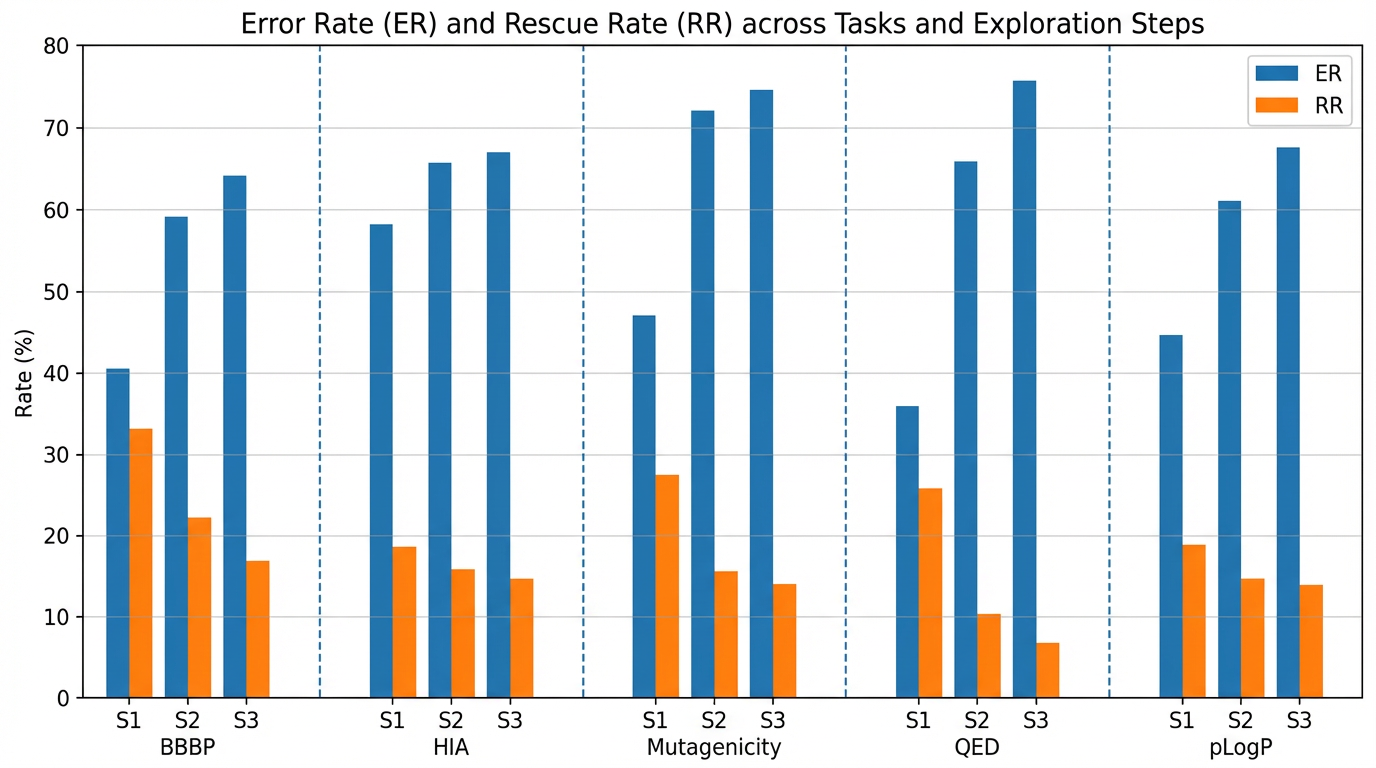}

\caption{In-context self-correction across \textsc{TRACE} exploration steps on five ADMET tasks, reported with Error Rate (ER) and Rescue Rate (RR).}
\label{fig:self_correction}

\end{figure}

\vspace{+0.05in}
\paragraph{System Configuration and Agent Variants.} 
TRACE instantiates four lead optimization tools, each corresponding to a GeLLMo
variant finetuned on MuMOInstruct but differing in backbone and the size of the
chemical property power set used during training. Specifically, we use
GeLLMo-P(4)-Llama, GeLLMo-P(6)-Llama, GeLLMo-P(4)-Mistral, and
GeLLMo-P(6)-Mistral, which provide complementary strengths across ADMET tasks.
A GPT-4o--mini~\citep{openai_gpt4o_mini_docs} reasoner selects tools and prompt
templates based on tool descriptions and the current context, and the orchestrator
executes the resulting plan with evolutionary exploration. Candidates are evaluated
using RDKit~\citep{rdkit} for \textsc{pLogP} and \textsc{QED}, and ADMET-AI~\citep{swanson2024admet}
for \textsc{BBBP}, \textsc{HIA}, and \textsc{Mutagenicity}.

\begin{table}[htbp]
\caption{Ablation on evolutionary exploration. We report \textsc{BF} (BestFrom) and \textsc{Nov} (Novelty) across exploration steps.}

\label{table:evolution}
\centering
\scriptsize
\setlength{\tabcolsep}{3pt}

\begin{tabular}{l*{10}{c}}
\hline
\textbf{Method} &
\multicolumn{2}{c}{\textbf{bbbp}} &
\multicolumn{2}{c}{\textbf{hia}} &
\multicolumn{2}{c}{\textbf{mutagenicity}} &
\multicolumn{2}{c}{\textbf{qed}} &
\multicolumn{2}{c}{\textbf{plogp}} \\
\cline{2-11}

& BF & Nov
& BF & Nov
& BF & Nov
& BF & Nov
& BF & Nov \\
\hline

TRACE-Parallel-S1 & 5.51 & 46.59 & 10.87 & 36.66 & 11.94 & 47.70 & 0.00 & 44.96 & 3.59  & 44.98 \\
TRACE-Parallel-S2 & 31.50 & 33.37 & 32.06 & 30.22 & 25.25 & 27.15 & 0.00 & 25.42 & 0.00 & 34.50 \\
TRACE-Parallel-S3 & 62.99 & 25.68 & 57.07 & 25.63 & 62.80 & 23.54 & 100.00 & 16.05 & 96.41 &  26.99 \\
\hline
\end{tabular}
\end{table}

We evaluate three TRACE variants that differ in per-step execution policy.
TRACE-Online executes one tool per step without trajectory reuse; TRACE-Retrieve
also executes one tool but performs similarity-guided trajectory retrieval to reuse tool sequences obtained from the training set; TRACE-Parallel is a high-budget upper bound that executes multiple tools and their prompt variants in parallel and selects the best candidate
anchored to the original lead. Unless stated otherwise, we run three exploration
steps. Additional implementation details are provided in
Appendix~\ref{app:impl}.

\vspace{+0.05in}
\paragraph{Evaluation Metrics.} We evaluate optimization performance using four metrics that capture
optimization success, structural fidelity, improvement magnitude, and chemical
validity~\citep{dey2025gellm3o}.
Success Rate (SR) defined as $\mathrm{SR}=\frac{\#\{\text{valid and correctly optimized samples}\}}{\#\{\text{all samples}\}}$, measures the fraction of samples for which the generated molecule
is chemically valid and successfully achieves better target property than the input
lead.
Similarity (SIM) quantifies structural fidelity by computing the average Morgan
fingerprint Tanimoto similarity between each input molecule $m_i$ and its optimized
output $m_i'$ over successful samples, $\mathrm{SIM}=\frac{1}{N_{\mathrm{succ}}}\sum_{i\in \mathrm{Succ}}\mathrm{Tanimoto}(m_i,m_i')$.
Relative Improvement (RI), defined as $\mathrm{RI}=\frac{1}{N_{\mathrm{ri}}}\sum_{i\in \mathrm{Succ},\,\mathrm{SIM}_i\ge 0.5}\frac{\Delta p_i}{p_i}$, measures the average normalized property gain among
successful samples that satisfy a similarity constraint where $p_i$ is the initial property value and $\Delta p_i$ is its absolute change.
Validity Rate (VR) reports the fraction of generated molecules that correspond to
valid SMILES strings, $\mathrm{VR}=\frac{\#\{\text{valid SMILES}\}}{\#\{\text{all generated samples}\}}$.These metrics assess whether a method produces chemically valid molecules, achieves substantive improvement, and preserves structural integrity during
optimization.

\vspace{+0.05in}
\paragraph{Dataset.} We conduct experiments on the MuMOInstruct dataset~\citep{dey2025gellm3o}. The dataset comprises around 255k molecule pairs, where each pair contains a source molecule and a derived counterpart obtained through a minimal and chemically interpretable modification, such as functional-group substitution or small substructure replacement that improves the target property without substantially altering the underlying molecular scaffold. By design, the modifications are chemically interpretable (e.g., adding a hydroxyl
group to improve solubility) and yield molecules that remain recognizably similar to the original lead.
This construction allows the dataset to simulate realistic lead optimization
scenarios, where a plausible improvement is known to exist through small, structure-preserving changes. We follow the official train/test splits, thereby providing a fair evaluation of the agent's ability to discover effective optimization trajectories at inference time.

\section{Experimental Results}

\vspace{+0.05in}
\paragraph{Main Results.} 
Table~\ref{table:main} reports results across five ADMET-related lead
optimization tasks. Overall, \textsc{TRACE} variants perform best across model
families, achieving higher success rates and validity with larger relative
improvements while maintaining sufficient structural similarity.

GraphGA~\citep{jensen2019graph} provides a classical non-LLM reference, but it is
only applicable to the \textsc{pLogP} objective in our setup and does not
naturally enforce lead-conditioned similarity constraints because it relies on
population-level mutation and selection. Consequently, it achieves perfect
validity but extremely low similarity. Moving to LLM-based baselines,
general-purpose LLMs are not competitive, exhibiting low success and modest
improvements. Most chemistry-oriented foundation LLMs are stronger but often trade
success for structural fidelity, suggesting larger edits to the input lead. In
contrast, task-specific lead optimization LLMs offer the most competitive
single-model baseline with higher similarity and more reliable improvements, yet
they still fall short of \textsc{TRACE} in both success and improvement magnitude.

Across all tasks, \textsc{TRACE} achieves near-saturated success and perfect
validity under explicit similarity constraints, and its larger relative gains
highlight the benefit of multi-step, tool-orchestrated exploration. Comparing
variants reveals a budget--performance trade-off: \textsc{TRACE-Parallel} acts as
a high-budget upper bound, while \textsc{TRACE-Online} remains strong with a
single tool call per step and \textsc{TRACE-Retrieve} further improves success
and gains via similarity-guided trajectory reuse, recovering much of the benefit
of parallel exploration at a fraction of the compute cost. Similarity remains
above the predefined threshold across variants, indicating scaffold-preserving
optimization throughout.

\vspace{+0.05in}
\paragraph{Ablation Studies on Self-Correction.} 
Figure~\ref{fig:self_correction} evaluates in-context self-correction under
increasing exploration depth using Error Rate (ER) and Rescue Rate (RR) across
five ADMET tasks. As depth increases, optimization moves further from the initial
lead and failures become more frequent, reflected by a consistent rise in ER
across properties. Despite the harder recovery regime, self-correction remains
effective: although RR decreases with depth, it still rescues a substantial
fraction of failed generations across tasks and exploration steps. Overall, these results
show that in-context self-correction consistently mitigates failures throughout
multi-step search by leveraging intermediate outcomes as context.

\vspace{+0.05in}
\paragraph{Ablation Studies on Exploration.} 
We study the effect of exploration depth by measuring (i) when the best molecule
along a trajectory is found and (ii) how many successful and novel candidates are introduced at each step.
Table~\ref{table:evolution} reports \textsc{BF} (BestFrom), the fraction of test
cases whose best-performing molecule is obtained at a given step, and
\textsc{Nov} (Novelty), the fraction of successful candidates that are novel
relative to all molecules generated in earlier steps. Across tasks, \textsc{BF}
shifts toward deeper steps, indicating that many high-quality solutions emerge
after additional steps of exploration. Meanwhile, \textsc{Nov} remains
substantial even at later steps, showing that deeper exploration continues to
discover new constraint-satisfying candidates.

\section{Case Study}
\label{sec:case_study}

\begin{figure*}[!ht]
  \centering
  \includegraphics[width=0.95\linewidth]{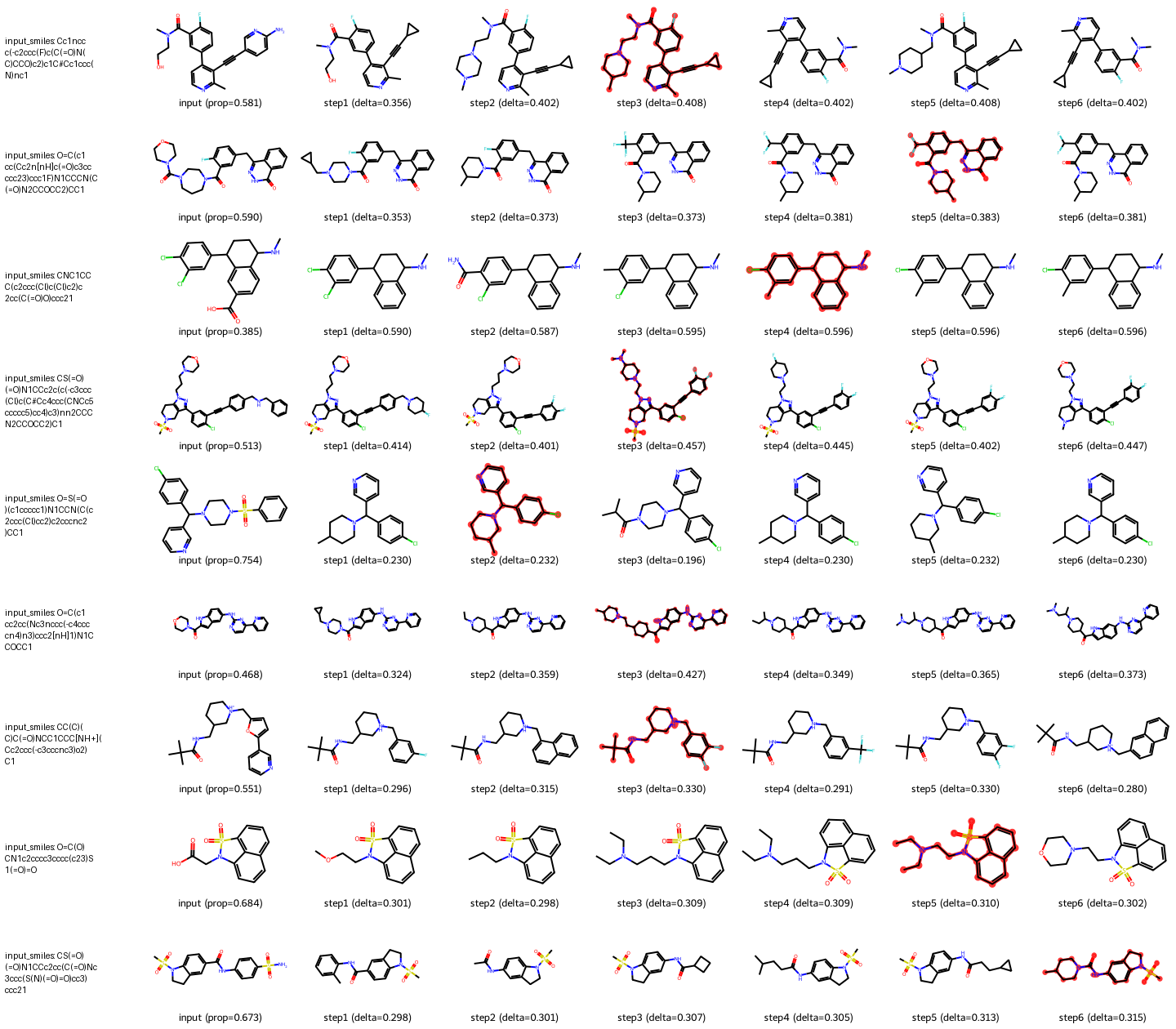}
  \caption{Representative \textsc{BBBP} optimization trajectories produced by \textsc{TRACE}. Each row starts from an input lead and shows molecules generated across exploration steps; the highlighted molecule achieves the largest relative improvement (\texttt{delta}) along the trajectory, measured w.r.t.\ the original lead (\texttt{prop} denotes the predicted BBBP score).}
  \label{fig:case_study_bbbp}
\end{figure*}

Figure~\ref{fig:case_study_bbbp} visualizes representative \textsc{BBBP}
optimization trajectories produced by \textsc{TRACE}. Each row corresponds to
a single lead molecule, and each column is a successive exploration step; we
annotate each step with \(\Delta\), the property gain relative to the input
lead, and highlight in red the molecule that achieves the best property value
along the trajectory. Across all cases, \textsc{TRACE} consistently preserves
the dominant scaffold and performs targeted edits on peripheral functional
groups. This is the regime in which medicinal chemistry expects meaningful
ADMET changes with minimal disruption to the presumed pharmacophore.

\medskip
\noindent\textbf{Case~1: Scaffold preservation and peripheral acid removal.}\\
{\footnotesize\texttt{Lead SMILES: CNC1CCC(c2ccc(Cl)c(Cl)c2)c2cc(C(=O)O)ccc21}}\\[2pt]
The starting molecule contains a fused bicyclic scaffold bearing a carboxylic
acid substituent and dichloro-aryl decoration. Across the trajectory the fused
ring system remains intact. \textsc{TRACE}'s first move is a functional-group
substitution that removes the peripheral carboxylic acid; subsequent steps
briefly test a carboxamide variant and then perform a small substituent walk
on the left aromatic ring, adjusting the chlorine substituents toward a small
hydrocarbon group. The best molecule preserves the same core and overall
topology while differing only in peripheral functional groups, reducing the
polarity liability and halogen burden. This trajectory exemplifies anchored
multi-step refinement: \textsc{TRACE} makes one decisive functional-group
modification, then fine-tunes substituents without destabilizing the scaffold.
The edits are interpretable, minimal, and chemically plausible, the kind of
scaffold-preserving micro-optimization that often yields large ADMET gains
without altering scaffold connectivity.

\medskip
\noindent\textbf{Case~2: Peripheral handle redesign via acid\(\rightarrow\)amine remodeling.}\\
{\footnotesize\texttt{Lead SMILES: O=C(O)CN1c2cccc3cccc(c23)S1(=O)=O}}\\[2pt]
Here the lead contains a sulfonamide-linked polyaromatic core and a terminal
carboxylic acid. \textsc{TRACE} progressively rewires the peripheral handle
while keeping the sulfonamide--aromatic fragment fixed. The turning point is
replacement of the terminal acid with an amine-containing chain, maintaining
the core scaffold while modifying the terminal substituent. \textsc{TRACE}
then explores alternative amine variants, including both acyclic and cyclic
configurations such as a morpholine-like ring that raises polarity and
H-bond acceptor count, before converging on a compact tertiary amine. This
focused side-chain walk preserves the central scaffold and proceeds through
systematic redesign of the peripheral substituent that most directly controls
charge state, polarity, and distribution.

\medskip
\noindent\textbf{Case~3: Conservative local edits followed by a bolder, higher-reward redesign.}\\
{\footnotesize\texttt{Lead SMILES: O=C(c1cc2cc(Nc3nccc(-c4ccccn4)n3)ccc2[nH]1)N1CCOCC1}}\\[2pt]
This lead contains a flexible, heterocycle-rich scaffold. \textsc{TRACE}
begins with conservative changes at the extreme periphery, introducing small
alkyl substitutions (cyclopropyl- or ethyl-like groups) while leaving the
central framework unchanged. The best-performing molecule appears
mid-trajectory, where \textsc{TRACE} introduces a more substantial peripheral
reconfiguration involving changes in ring or heteroatom placement while
maintaining overall scaffold connectivity and topology. This
``small-first, bold-later'' trajectory is consistent with \textsc{TRACE}
behaving as a budget-aware planner: it gathers signal with low-risk edits,
then attempts a higher-reward modification once the context suggests a
profitable direction. The non-monotonic improvement also motivates
\textsc{TRACE}'s best-seen selection, which explicitly tracks and returns
the best candidate observed along the trajectory rather than assuming
steady progress.

\paragraph{Takeaway.} Across these cases, \textsc{TRACE}'s strength lies not
in a single ``magic transformation'' but in a consistent structural
refinement pattern: (i) preserve the core scaffold, (ii) modify high-leverage
peripheral functional groups, and (iii) iteratively explore localized
alternatives under similarity constraints. The best-performing molecules
often emerge mid-trajectory rather than at the last step, supporting
\textsc{TRACE}'s design choice to track and return the best-seen candidate
during multi-step exploration instead of assuming monotonic improvement.

\section{Conclusion}

We propose TRACE, a trajectory-aware, multi-tool agentic framework for molecular lead optimization that treats optimization as a standalone decision process rather than a single execution step. By combining adaptive tool orchestration, in-context self-correction, and anchored multi-step evolutionary exploration, \textsc{TRACE} coordinates heterogeneous instruction-guided optimizers to progressively refine candidates while preserving key molecular substructures under explicit similarity constraints.
To operate under limited resources, \textsc{TRACE} introduces similarity-guided trajectory reuse, allowing the agent to recover much of the benefit of parallel multi-tool exploration at a fraction of the computational cost. Extensive
experiments across multiple ADMET tasks demonstrate consistent gains in validity, success, and improvement magnitude, while ablations confirm the roles of
self-correction, exploration depth, and trajectory reuse in stabilizing and improving optimization. Together, these results highlight the potential of trajectory-aware agentic optimization as a practical and effective approach for lead optimization in realistic drug design workflows.

\section{Limitations and Ethical Considerations}
\label{sec:limitations_ethics}

\paragraph{Limitations.}
Our study focuses on lead-conditioned optimization of a set of ADMET-related
objectives under explicit similarity constraints, using predictive evaluators.
Future work can extend \textsc{TRACE} by incorporating target-specific objectives and practical developability constraints such as synthetic accessibility.

\paragraph{Ethical considerations.}
We do not introduce new ethical concerns beyond those common to molecular design
and optimization tools, and our study uses no human-subject data. We recommend
standard safeguards before downstream use.

\paragraph{Data privacy and consent.}
Our experiments use publicly available datasets and do not involve private or
human-subject data.

\section*{Acknowledgements}

This material is based in part upon work supported by the National Science
Foundation under Grant IIS-2212174, National Institute of Aging (NIA)
R01AG072449, and National Institute of General Medical Sciences (NIGMS)
R01GM145700.

\bibliographystyle{ACM-Reference-Format}
\bibliography{reference}

\clearpage
\appendix

\section{Framework Components}
\label{app:frame}

\subsection{LLM Reasoner}
\label{app:frame_llm-reasoner}

The LLM reasoner serves as the high-level reasoning and decision-making component
within the overall agentic framework.
At each decision point, the reasoner receives a structured context constructed by the orchestrator, which includes tool descriptions and trajectory-level experience.
Based on this context, it determines the next action to take, such as selecting
an optimization tool, refining execution instructions, or advancing the current
optimization trajectory.

Importantly, the reasoner does not directly generate optimized molecular
structures.
Instead, it produces structured commands or prompts that specify how downstream
tools should be invoked.
This design decouples reasoning from execution, allowing the LLM to focus on
strategic decision making while delegating domain-specific molecular operations
to specialized tools.
Because decisions are conditioned on an evolving context, the reasoner can adapt
its behavior dynamically as new outcomes are observed, without requiring any
parameter updates or task-specific retraining.

\subsection{Orchestrator}
\label{app:frame_orchestrator}

The orchestrator acts as the central control module of the framework and mediates
all interactions among components.
It maintains the global execution state and ensures coherent coordination between
the LLM agent, generative tools, predictive evaluators, and memory modules.
Through centralized control, the orchestrator enforces consistent execution
semantics while enabling adaptive behavior driven by feedback.

Concretely, the orchestrator performs three core functions.
First, it retrieves relevant information from tool instruction memory and the trajectory buffer to assemble the contextual input provided to the LLM agent.
Second, it dispatches generative tools according to the agent’s decisions and
manages their execution, including input preparation and output collection.
Third, it performs instruction and context-level refinement by updating prompts and constraints in response to observed failures or suboptimal outcomes.
This feedback-driven refinement allows the system to correct ineffective behaviors and stabilize execution over time.

\subsection{Tool Specifications and Result History}
\label{app:frame_memory}

Our framework maintains two complementary sources of system knowledge:
(i) a set of explicit tool specifications that define how domain-specific tools
are invoked, and
(ii) a result history that records what has been generated and evaluated during
optimization.
Together, these components enable the agent to reason over both what tools
are capable of and what has already occurred in the optimization process.

Tool specifications provide structured descriptions of available
domain-specific tools, including their expected inputs, operational constraints, and predefined prompt templates.
They serve as a stable interface between the LLM agent and heterogeneous molecular generators, ensuring that tool selection and invocation remain aligned with chemical domain knowledge.
By externalizing tool definitions from the agent itself, the framework can easily incorporate new tools or update existing ones without modifying the agent architecture or retraining the agent.

Result history records molecular candidates generated during execution,
together with their predictive evaluation scores and relevant metadata.
This history captures factual information about past outcomes, including which
candidates improved target properties and which failed to satisfy structural, property, or validity constraints. By explicitly retaining past results, the framework enables cumulative improvement and avoids redundant exploration of previously ineffective regions of chemical space.

\subsection{Trajectory Buffer}
\label{app:frame_trajectory-buffer}

The trajectory buffer stores historically successful lead optimization
trajectories collected from optimization runs on the training dataset.
Each trajectory consists of an ordered sequence of intermediate molecules and
the corresponding tools applied at each step, capturing how a lead compound was
progressively refined over multiple tool invocations. 
It encodes how different tools can be sequenced to achieve gradual property
improvement while preserving key structural motifs.
These trajectories reflect accumulated domain knowledge about effective
tool-usage strategies under structural constraints.

During inference, the buffer supports similarity-guided trajectory retrieval.
Given a current molecule, the agent searches for structurally similar molecules in the buffer and retrieves the corresponding high-performing optimization trajectories.
Retrieved trajectories are not executed verbatim.
Instead, they are used as soft guidance for tool sequencing, either by initializing a candidate tool plan or by adjusting an ongoing optimization trajectory.
This retrieval-based mechanism allows the agent to reuse prior optimization
experience while remaining flexible to adapt decisions based on newly observed
outcomes.

\subsection{Lead Optimization Tools}
\label{app:frame_tools}

The lead optimization tools in our framework are LLM-based models that operate directly through natural-language instructions.
Given a lead molecule and a textual prompt specifying optimization objectives or constraints, each tool proposes modified molecular candidates aimed at improving target properties while preserving key structural features.
Such instruction-guided LLM models have demonstrated strong performance in recent work on lead optimization tasks.

These tools leverage the in-context reasoning capabilities of LLMs to interpret complex optimization instructions and adapt their outputs accordingly—an ability that is absent in traditional, task-specific lead optimization models.
Within the framework, each tool functions as an independent optimization module, and its internal generation process remains unchanged.
The overall system coordinates multiple heterogeneous lead optimization tools by
selecting, sequencing, and reusing them based on observed behaviors and outcomes, while maintaining modularity and extensibility as new tools become available.

\subsection{Predictive Evaluators}
\label{app:frame_predictors}

Predictive evaluators provide quantitative assessments of generated molecular candidates with respect to target properties and structural constraints. They include existing, pre-trained property predictors and chemical checks, such as ADMET-related scores, validity verification, and structural similarity measures. Evaluation results are fed back to the orchestrator and recorded in the result
history. They serve as objective signals for candidate comparison and prompt refinement. Because evaluators are modular and decoupled from the agent and lead optimization tools, additional property objectives can be incorporated by simply adding the corresponding predictors. This design allows the agent to adapt to evolving optimization requirements without retraining and ensures that optimization behavior remains grounded in consistent, quantitative performance signals.

\section{Implementation Details}
\label{app:impl}

\subsection{Lead Optimization Tools}
\label{app:impl_tools}
We instantiate four instruction-conditioned lead-optimization tools, each
implemented as a distinct GeLLMo checkpoint. All tools are instruction-finetuned
on MuMOInstruct, but differ in (a) the base language backbone and (b) the size of
the chemical property power set used during training. We follow~\citep{dey2025gellm3o} to denote variants as
GeLLMo-P($k$)-X, where $k\in\{4,6\}$ controls the property power-set size and
X$\in\{$Llama, Mistral$\}$ denotes the language backbone.

\subsection{Rationale for Tool Selection}
\label{app:impl_tools_selection}

Our framework focuses on lead-conditioned optimization of ADMET-related properties
under explicit structural constraints. Prior molecule optimization methods such as
MolDQN, JT-VAE, and GraphGA are commonly evaluated under different settings: they
often target relatively simple objectives, may not operate through localized
motif-level substitutions, and frequently do not enforce per-lead similarity
constraints, which can lead to substantial structural drift. These mismatches make
them less aligned with our localized, constraint-aware lead refinement setting~\citep{jin2018junction, zhou2019optimization, jensen2019graph}.

\subsection{Computational Cost}
\label{app:impl_cost}

We report computational cost under two execution budgets. For TRACE-Online
and TRACE-Retrieve, we run inference on a single NVIDIA RTX A5000 GPU,
and the average optimization time is under 1 second per molecule per stage.
In contrast, TRACE-Parallel executes all tools in parallel and therefore
requires four GPUs for simultaneous inference; overall, its compute cost is
approximately $24\times$ the low-budget setting.

\subsection{Reasoner System Prompt Template}
\label{app:impl_template}

\vspace{0.25em}
\noindent\fbox{%
\begin{minipage}{0.98\linewidth}
\ttfamily\footnotesize
You are a planner (not a generator) for constrained lead optimization.\\
Given an input SMILES and a target property, select tool calls under the current
budget policy.\\
\textbf{Online/Retrieve:} output an \emph{ordered} sequence of tool calls.\\
\textbf{Parallel:} output a \emph{set} of tool calls to be executed in parallel.\\
For each selected tool, choose one prompt\_index in Online/Retrieve.
In parallel mode, you may choose multiple indices for the same tool.\\[0.5em]

Return ONLY a JSON object that follows this schema:\\
\{"tool\_calls":[\{"tool\_name":"<string>", "prompt\_index":<0--5>\}, ...]\}\\[0.5em]

Input SMILES: \{input\_smiles\}\\
Target property: \{target\_property\}\\
Retrieval hint (TRACE-Retrieve only): \{retrieval\_hint\}\\[0.5em]

Available tools:\\
\{tool\_specs\}\\[0.5em]

Prompt templates (indexed 0--5):\\
\{prompt\_templates\}
\end{minipage}}

\end{document}